\documentclass{amia}
\usepackage{graphicx}
\usepackage[labelfont=bf]{caption}
\usepackage[superscript,nomove]{cite}
\usepackage{color}
\usepackage{url}
\usepackage{amsmath}
\usepackage{amsfonts}
\usepackage{amssymb}
\usepackage{booktabs}
\usepackage{wrapfig}
\usepackage{ulem}
\usepackage{tabularx}
\usepackage{enumitem}

\begin{document}

\title{Using Radiomics as Prior Knowledge for Thorax Disease Classification and Localization in Chest X-rays}

\author{Yan Han$^1$, Chongyan Chen$^1$, Liyan Tang$^1$, Mingquan Lin$^{2}$, Ajay Jaiswal$^1$, Song Wang$^1$, Ahmed Tewfik$^1$, George Shih$^3$, Ying Ding$^{1,*}$, Yifan Peng$^{2,}$\footnote{Equal contributions.}}

\institutes{$^{1}$The University of Texas at Austin, Austin, TX, USA; $^{2}$ Population Health Sciences, Weill Cornell Medicine, New York, NY, USA; $^{3}$ Department of Radiology, Weill Cornell Medicine, New York, NY, USA; 
}

\maketitle

\noindent{\bf Abstract}

\textit{Chest X-ray becomes one of the most common medical diagnoses due to its noninvasiveness. The number of chest X-ray images has skyrocketed, but reading chest X-rays still have been manually performed by radiologists, which creates huge burnouts and delays. Traditionally, radiomics, as a subfield of radiology that can extract a large number of quantitative features from medical images, demonstrates its potential to facilitate medical imaging diagnosis before the deep learning era. In this paper, we develop an end-to-end framework, ChexRadiNet, that can utilize the radiomics features to improve the abnormality classification performance. Specifically, ChexRadiNet first applies a light-weight but efficient triplet-attention mechanism to classify the chest X-rays and highlight the abnormal regions. Then it uses the generated class activation map to extract radiomic features, which further guides our model to learn more robust image features. After a number of iterations and with the help of radiomic features, our framework can converge to more accurate image regions. We evaluate the ChexRadiNet framework using three public datasets: NIH ChestX-ray, CheXpert, and MIMIC-CXR. We find that ChexRadiNet outperforms the state-of-the-art on both disease detection (0.843 in AUC) and localization (0.679 in T(IoU) = 0.1). We will make the code publicly available at \url{https://github.com/bionlplab/lung_disease_detection_amia2021}, with the hope that this method can facilitate the development of automatic systems with a higher-level understanding of the radiological world.}

\section{Introduction}
The chest X-ray is one of the most common medical procedures for diagnosis, but the interpretation of chest x-ray images is subject to significant diagnosis variability for important clinical decisions. A radiologist reads about 20,000 images a year, roughly 50-100 per day, and the number is increasing. Each year, the US produces 600 billion images, and 31\% of American radiologists have experienced at least one malpractice claim, often missed diagnoses\cite{topol2019deep}. The shortage of radiologists and burnout of physicians creates an urgent demand for immediate solutions. Building automatic or semi-automatic approaches to medical imaging diagnosis becomes an unavoidable next step.

The recent development of artificial intelligence, especially deep learning, offers great potential to improve medical imaging diagnosis\cite{ching2018opportunities}. It also sneaks into the radiology reading rooms to build a new paradigm for precision diagnosis\cite{rajpurkar2017chexnet,wang2017chestxray8,wang2018tienet}. Pioneering work on chest X-rays mainly focused on two problems: disease classification and localization. The recent release of large-scale datasets, such as NIH Chest X-ray\cite{wang2017chestxray8}, CheXpert\cite{irvin2019chexperta}, and MIMIC-CXR\cite{johnson2019mimiccxrjpg}, have enabled many studies using deep learning for automated chest X-ray diagnosis, such as thorax disease classification~\cite{rajpurkar2017chexnet,sowrirajan2020moco,yao2017learning,gundel2019learning} and localization~\cite{wang2017chestxray8,li2018thoracic,hwang2016self}.

%Most existing work on disease detection and localization on chest X-ray are data-driven. 
In practice, radiologists use pattern recognition on medical images to make a diagnostic decision\cite{bryan2001introduction}. The knowledge of radiologists can be captured by \uline{Radiomics}, which has demonstrated the effectiveness of image-based biomarkers for cancer staging and prognostication. Formally, radiomics extracts quantitative data from medical images to represent tumor phenotypes, such as spatial heterogeneity of a tumor and spatial response variations. It plays an important role in precision medicine to support evidence-based clinical decision-making. For example, radiomics can generate the detailed quantification of tumor phenotype\cite{nicolasjilwan2015addition} and acts as a radiographic imaging phenotype which is associated with tumor stage, metabolism, and gene or protein expression profiles\cite{ganeshan2010texture,ganeshan2013non}. 
%Eilaghi et al. studied that radiomics of CT texture features of the dissimilarity are associated with the overall survival of pancreatic cancer\cite{eilaghi2017ct}. Chen et al. showed that the first-order radiomic features (e.g., mean, skewness, and kurtosis) are correlated with pathological responses to cancer treatment\cite{chen2017assessment}. Huang et al. showed that radiomics could increase the positive predictive value and reduce the false-positive rate in lung cancer screening for small nodules compared with a human reading by thoracic radiologists\cite{huang2018addeda}. Zhang et al. found that multiparametric MRI-based radiomics nomograms provided improved prognostic ability in advanced nasopharyngeal carcinoma (NPC)\cite{zhang2018learning}.

While radiomics offer the potential for more precise and accurate clinical predictions, it is surprising that radiomics has not been implemented in the layers of the neural networks, nor to the best of our knowledge in the deep learning workflow for X-ray analysis\cite{parekh2019deep,liu2019clinically}. To bridge this gap, in this paper, we propose ChexRadiNet, a new framework that incorporates domain-specific knowledge (radiomics) into deep learning algorithms as soft constraints, and then learns end-to-end to automatically detect thorax diseases and generate bounding boxes on chest X-rays. Compared with previous studies, our proposed model does not need pre-annotated bounding boxes for training and can achieve state-of-the-art performance for thorax disease localization. Therefore, it provides a way to introduce prior information about anticipated explanations, a technique that is widely used in the “Rationale model”\cite{lei2016rationalizinga} (Section~\ref{sec:method}).
For ensuring ChexRadiNet is robust and generalizable, three public benchmarking datasets were used for this purpose: NIH Chest X-ray\cite{wang2017chestxray8}, CheXpert\cite{irvin2019chexperta}, and MIMIC-CXR\cite{johnson2019mimiccxrjpg}. We demonstrate that our model outperforms baseline methods for both thorax disease classification and localization (Section~\ref{sec:exp}).

\section{Method}
\label{sec:method}

\begin{figure}[t]
\centering
\includegraphics[width=.8\textwidth,clip,trim=0 20em 20em 0]{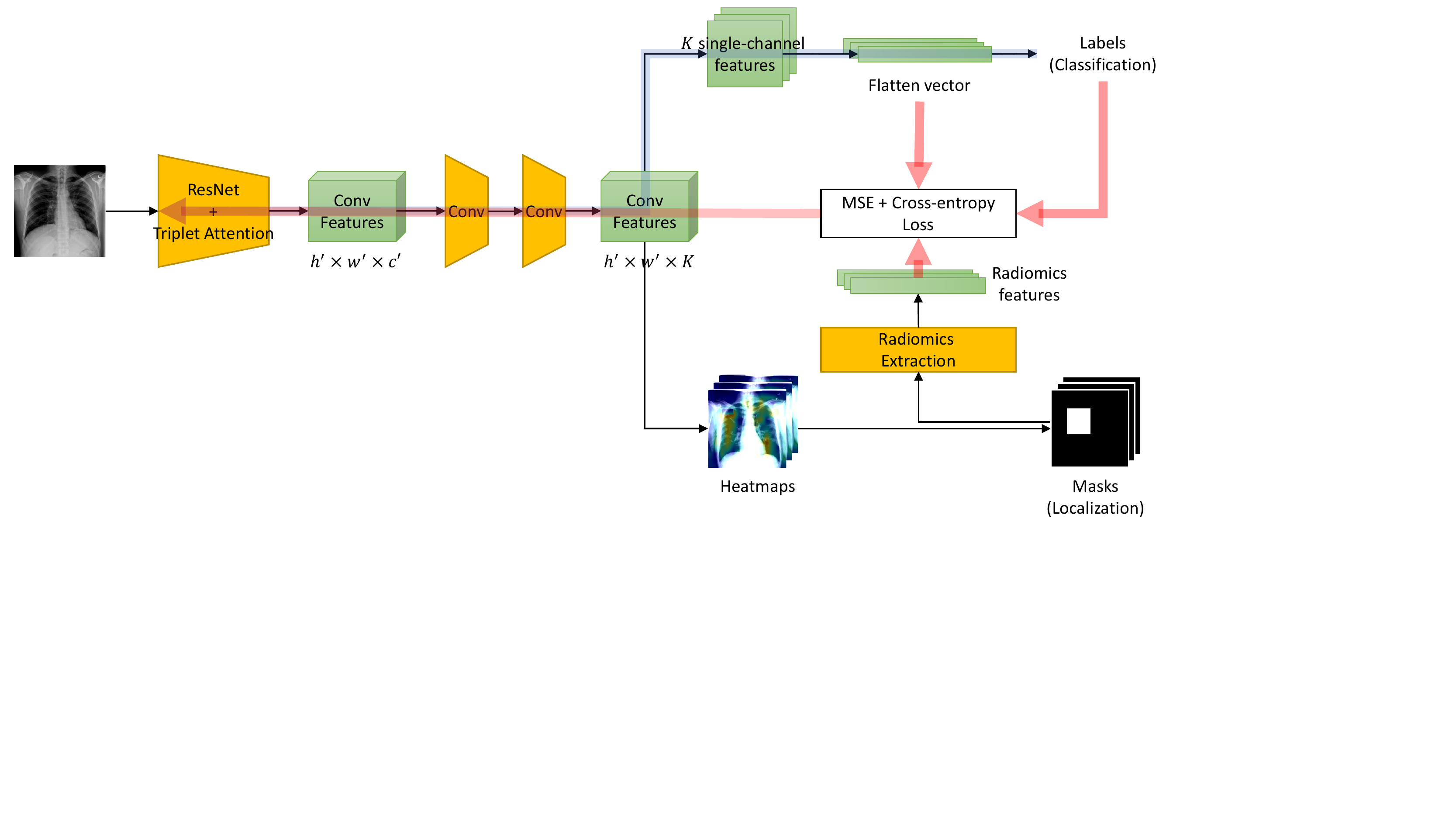}
\vspace{-1em}
\caption{Model overview. The model contains three major parts. Blue arrows represents the feedforward multi-label classification part. The below black arrows represents the mask generation and radiomic features extraction part. Red arrows means the radiomic features regularization and backward part.}
\label{fig:model}
\end{figure}

Figure~\ref{fig:model} shows our proposed ChexRadiNet, which consists of two branches. The first branch predicts whether the pathology is present or not in the image. The second branch localizes its regions using the radiomic features extracted from the first branch. ChexRadiNet utilizes a multi-task, closed-loop strategy to learn and use radiomic features as soft constraints. Formally, we are learning a two-part latent-variable model of the form $E_{z\sim p(z|x)}p(y|x,z)$, where the latent $z$ is a radiomic-based mask over the image $x$ with the probability $p(z|x)$. $p(y|x,z)$ is a masked version of the classification framework. Therefore, we consider the training process as a weakly-supervised learning. In this section, we first illustrate the architecture of ChexRadiNet and then present the training process.

\subsection{Model architecture}

\subsubsection{Branch I: Multi-label classification}

In this branch, we label each image with a 14-dim vector $y=[y_1,\ldots,y_k,\ldots,y_K]$, $y_k\in {0,1}$, $K=14$ for each image. $y_k$ indicates the presence with respect to the according pathology in the image while a zero vector represents the status of “Normal” (no pathology is found in the scope of any of 14 disease categories as listed).

We use the residual neural network (ResNet) architecture\cite{he2016deep}, given its dominant performance in ILSVRC competitions and the triplet attention mechanism (see Section~\ref{sec:triplet}). However, our framework can be applied to other CNNs. ResNet-18 and ResNet-50 are used in this paper. After removing the final classification layer and global pooling layer, an input image with shape $h\times w\times c$ produces a feature tensor with shape $h'\times w'\times c'$ where h, w, and c are the height, width, and number of channels of the input image, respectively while $h'=h/32$,  $w'=w/32$, $c'=2048$. The output of this network encodes the images into a set of abstracted feature maps. Then through an application of two convolutional layers (each followed by batch normalization and ReLU activation), the number of channels is modified to $K$, where $K$ is the number of possible disease types. A perchannel probability for each disease class is then derived by a fully-connected layer with a sigmoid activation function; this is denoted $p(k|I)$, where the probability is that whether the image belongs to class $k$ and $I$ denotes the image. Since we intend to build $K$ binary classifiers, we will exemplify just one class $k$. Note that $k$th binary classifiers will use the $k$th-channel features to do prediction. Since all images have their labels, the loss function for class $k$ can be expressed as minimizing the binary-cross entropy as $L_k=-y_k\log p(k|I)-(1-y_k)\log (1-p(k|I))$, where $y_k$ is the ground truth label of the $k$ class. To enable end-to-end training across all classes, we sum up the class-wise losses to define the total loss as $L_{I}=\sum_k{L_k}$.

\subsubsection{Branch II: Mask generation}
In this branch, we generate bounding boxes (B-Box, or masks) based on the classification result of Branch I to get the most indicative areas using the class activation mappings (CAMs)\cite{zhou2016learninga}. The heatmap produced from the model indicates the approximate spatial location of one particular thoracic disease class each time. Due to the simplicity of intensity distributions in these resulting heatmaps, applying an ad-hoc thresholding-based B-Box generation method for this task is found to be sufficient. Followed by the work of Wang et al.\cite{wang2017chestxray8}, the intensities in heatmaps are first normalized to $[0, 255]$ and then thresholded by $\{60, 180\}$ individually. Finally, B-Boxes are generated to cover the isolated regions in the resulting binary maps. 

\textbf{Radiomic features extraction.} With the generated B-Boxes and original images, we extracted radiomic features to regularize the model. Quantitative radiomics can be categorized into the following subgroups: 
\vspace{-1em}
\setlist[itemize]{leftmargin=1em}
\begin{itemize}
\setlength\itemsep{0em}
\item First-order statistics features describe the distribution of individual pixel values without concerns for spatial relationships. They are histogram-based properties using mean, median, maximum, and minimum values of the pixel intensities on the image, as well as their asymmetry, flatness, uniformity, and entropy. 
\item Shape features describe the shape of the region of interest (ROI) and its geometric properties (e.g., volume, maximum diameter along with different orthogonal directions, maximum surface, tumor compactness, and sphericity). 
\item A Gray Level Co-occurrence Matrix (GLCM) features describe the second-order joint probability function of an image region constrained by the mask. The matrix $P(i,j|\delta,\theta)$ represents the number of times the combination of levels $i$ and $j$ occurs in two pixels in the image, that are separated by a distance of $\delta$ pixels along angle $\theta$. %The distance $\delta$ from the center pixel is defined as the distance according to the infinity norm. For $\theta=1$, this results in 2 neighbors for each of 13 angles in 3D (26-connectivity) and for $\theta=2$, a 98-connectivity (49 unique angles).
%\item A Gray Level Co-occurrence Matrix (GLCM) of size $N_g \times N_g$ describes the second-order joint probability function of an image region constrained by the mask and is defined as $P(i,j|\delta,\theta)$. The $(i,j)^{th}$ element of this matrix represents the number of times the combination of levels $i$ and $j$ occur in two pixels in the image, that are separated by a distance of $\delta$ pixels along angle $\theta$. The distance $\delta$ from the center pixel is defined as the distance according to the infinity norm. For $\theta=1$, this results in 2 neighbors for each of 13 angles in 3D (26-connectivity) and for $\theta=2$, a 98-connectivity (49 unique angles).
\item A Gray Level Size Zone (GLSZM) features quantify gray level zones in an image. A gray level zone is defined as the number of connected pixels that share the same gray level intensity. %A pixel is considered connected if the distance is 1 according to the infinity norm (26-connected region in a 3D, 8-connected region in 2D). In a gray level size zone matrix $P(i,j)$ the $(i,j)^{th}$ element equals the number of zones with gray level $i$ and size $j$ appear in image. Contrary to GLCM and GLRLM, the GLSZM is rotation independent, with only one matrix calculated for all directions in the ROI.
\item A Gray Level Run Length Matrix (GLRLM) features quantify gray level runs, which are defined as the length in number of pixels, of consecutive pixels that have the same gray level value. %In a gray level run length matrix $P(i,j)$, the $(i,j)^{th}$ element describes the number of runs with gray level $i$ and length $j$  occur in the image (ROI) along angle $\theta$.
\item A Neighboring Gray Tone Difference Matrix (NGTDM) features quantify the difference between a gray value and the average gray value of its neighbors within distance $\delta$. The sum of absolute differences for gray level $i$ is stored in the matrix. %Let $X_{gl}$ be a set of segmented pixels and  $x_{gl}(j_x,j_y,j_z)\in X_{gl}$ be the gray level of a pixel at position $(j_x,j_y,j_z)$, then the average gray level of the neigbourhood is 
%\[
%\frac{1}{W}\sum_{k_x=-\delta}^\delta{\sum_{k_y=-\delta}^\delta{\sum_{k_z=-\delta}^\delta{x_{gl}(j_x+k_x,j_y+k_y,j_z+k_z)}}}
%\]
%where $(k_x,k_y,k_z)\neq(0,0,0)$ and $x_{gl}(j_x+k_x,j_y+k_y,j_z+k_z)\in X_{gl}$. Here, $W$ is the number of pixels in the neighbourhood that are also in $X_{gl}$.
\item A Gray Level Dependence Matrix (GLDM) features quantify gray level dependencies in an image. A gray level dependency is defined as the number of connected pixels within distance $\delta$ that are dependent on the center pixel. %A neighbouring pixel with gray level $j$ is considered dependent on center pixel with gray level $i$ if $|i-j|\leq \alpha$. $\alpha$ is cutoff value for dependence. In a gray level dependence matrix  $P(i,j)$, the $(i,j)^{th}$ element describes the number of times a pixel with gray level $i$ with $j$ dependent pixels in its neighbourhood appears in image.
\end{itemize}

All above features can be extracted either directly from the images or after applying different filters or transforms (e.g., wavelet transform). In our design, we utilize the Pyradiomics tool to extract radiomic features (\url{https://pyradiomics.readthedocs.io/}).

Finally, we use the pairwise distance between radiomic features and image features as regularization. Therefore, the adjustable loss function is $L_{II}=L_I+ \| I_F-R_F \|_p$, where $I_F$ and $R_F$ are the image features and radiomic features, respectively, and $\|\cdot\|$  denotes the norm and $p$ represents the norm degree, e.g., $p = 1$ and $p = 2$  represent the Taxicab norm and Euclidean norm, respectively. In this paper, we set $p$ to 2. Please note that although the original shapes of $I_F$ and $R_F$ are not equal, we easily adapted one-layer MLP to project them into the same dimension space.

\subsubsection{Triplet Attention}
\label{sec:triplet}

To boost the quality of masks, we integrate the triplet-attention mechanism\cite{misra2020rotate}. Triplet Attention mechanism requires few learnable parameters and could capture important features by taking cross-dimension interaction into account\cite{misra2020rotate}. In other words, it includes three sub-branches to respectively capture the dependency between spatial dimensions Height ($H$), Width ($W$), and the Channel ($C$) dimension. For the first branch, in measuring the interactions between dimension $H$ and dimension $C$, it first performs a \textit{Z-pool} operation by concatenating the result of average pooling and max pooling across dimension $W$. This operation can be summarized as $\chi_1^*=\text{z-pool}(\chi')=[MaxPool_w (\chi');AvgPool_w(\chi')]$ where $\chi'\in \mathbb{R}^{W\times H\times C}$ is a 90 degree anti-clockwise rotation along the $H$ axis from the output of the previous convolutional layer $\chi\in \mathbb{R}^{C\times H\times W}$  and $\chi_1^*\in \mathbb{R}^{2\times H\times C}$ is  the  output  of  a  Z-Pool  operation. $\chi_1^*$ then passed through a standard 2D convolutional layer followed by sigmoid activation $\sigma$ to get attention weights for $\chi_1^*$. It would finally rotate back to match the original shape of $\chi$ after applying the attention weights. These steps can be represented by the following: $y_1=r(\chi'\sigma(\text{CNN}_1(\chi_1^*)))$  where $r$ is the rotation operation to retain the original shape of input. Similarly, $y_2$, $y_3$ are obtained from the last two branches by measuring the interactions between dimensions $W$ and $C$ and between dimensions $W$ and $H$, respectively. Note that the last branch is similar to the spatial attention in CBAM\cite{woo2018cbam}, and it requires no rotation. The refined input $y$ is represented by averaging outputs from three branches: $y=\frac{1}{3}(y_1+y_2+y_3)$.

\subsection{Training Strategy of ChexRadiNet}

ChexRadiNet adopts an end-to-end multi-task training scheme. Each epoch consists of two tasks. In the first task (Branch I), we use the whole image to fine-tune the ResNet + Triplet Attention network pre-trained on ImageNet. During this process, we feed the generated masks into the radiomics extraction block to get radiomic features. In the second task (Branch II), we use radiomic features as regularization to further fine-tune the whole model. In each epoch, we use the model with the highest AUC on the validation set for testing.

\section{Experiments}
\label{sec:exp}

\subsection{Datasets}

\begin{wraptable}{r}{6.5cm}
\centering
\caption{Descriptions of the datasets.}
\vspace{-1em}
\label{tab:dataset}
  \begin{tabular}{lrr}
\toprule
Datasets & Patients & Chest X-rays\\
\midrule
NIH Chest X-ray & 30,805 & 112,120\\
CheXpert & 65,240 & 224,316\\
MIMIC-CXR & 227,827 & 377,110\\
\bottomrule
 \end{tabular}
%\vspace{-1em}
\end{wraptable}

For the abnormality classification task, we evaluated the ChexRadiNet framework using the NIH Chest X-ray\cite{wang2017chestxray8}, CheXpert\cite{irvin2019chexperta}, and MIMIC-CXR\cite{johnson2019mimiccxrjpg} datasets (Table~\ref{tab:dataset}). The Chest X-ray dataset contains 112,120 X-ray images collected from 30,805 patients. The disease labels were extracted from radiological reports with Natural Language Processing tools\cite{peng2018negbio}. There are 15 classes, one for “No findings” and 14 diseases: Atelectasis, Cardiomegaly, Consolidation, Edema, Effusion, Emphysema, Fibrosis, Hernia, Infiltration, Mass, Nodule, Pleural thickening, Pneumonia, and Pneumothorax. The disease labels are expected to have above 90\% accuracy. In addition, the Chest X-ray dataset includes 984 bounding boxes for 8 types of chest diseases annotated for 880 images by radiologists. 

CheXpert dataset is another large-scale public chest X-ray dataset currently available, which contains 224,316 X-ray scans of 65,240 patients. This dataset was labeled for the presence of 14 observations, including 12 common thoracic pathologies. Each observation can be assigned to either positive (1), negative (0), or uncertain (-1). To simplify the task, we choose to ignore all the uncertain samples. In addition, to compare with previous literature, we follow the same evaluation protocol over 5 observations: Atelectasis, Cardiomegaly, Consolidation, Edema, and Pleural Effusion.
MIMIC-CXR is also a large-scale CXR dataset, which contains 377,110 chest X-rays associated with 227,827 imaging studies. Images are provided with 13 labels. Similar to CheXpert, each label can be assigned to either positive (1), negative (0), or uncertain (-1).

% \begin{table}[h]
% \centering
% \caption{Descriptions of the datasets.}
% \label{tab:dataset}
%   \begin{tabular}{lllp{15em}p{12em}}
% \toprule
% Datasets & Patients & Chest X-ray & Labels & B-Box\\
% \midrule
% NIH Chest X-ray & 30,805 & 112,120 & Atelectasis, Cardiomegaly, Consolidation, Edema, Effusion, Emphysema, Fibrosis, Hernia, Infiltration, Mass, Nodule, Pleural thickening, Pneumonia, and Pneumothorax & Atelectasis, Cardiomegaly, Effusion, Infiltration, Mass, Nodule, Pneumonia, and Pneumothorax\\
% \midrule
% CheXpert & 65,240 & 224,316 & Atelectasis, Cardiomegaly, Consolidation, Edema, Pleural Effusion & -\\
% \midrule
% MIMIC-CXR & 227,827 & 377,110 & Atelectasis, Cardiomegaly, Consolidation, Edema, Enlarged Cardiomediastinum, Fracture, Lung Lesion, Lung Opacity, Pleural Effusion, Pneumonia, Pneumothorax, Pleural Other, Support Devices & -\\
% \bottomrule
%  \end{tabular}
%\end{table}

\subsection{Evaluation metrics and experimental settings}
For the abnormality detection task, we randomly split each dataset into training (70\%), validation (10\%), and test (20\%) sets. Note that there is no patient overlap between the sets. We use AUC scores, the area under the ROC curve, to measure the disease identification accuracy. A higher AUC score indicates better performance.

For the abnormality localization task, following the work of Li et al\cite{li2018thoracic}, we only consider 8 diseases for the evaluation of mask generation because only eight types of diseases are provided with bounding boxes in the NIH Chest X-ray dataset. We use intersection over union (IoU) to evaluate the predicted disease regions against the ground truth bounding boxes.

We use ResNet-50 as the backbone model. We set the batch size as 256 and train the model for 20 epochs. The model is optimized using the stochastic gradient descent (SGD) optimizer with a learning rate of 0.1 and decay the learning rate by 0.1 every 5 epochs of training. We trained our model on AWS with 16 Nvidia K80 GPUs. The model is implemented in PyTorch.

\subsection{Results}

\subsubsection{Disease classification}

Table~\ref{tab:disease id} shows the AUC of each class and a mean AUC across the 14 chest diseases. We used ResNet-50 pre-trained on ImageNet as the backbone. Our ChexRadiNet outperforms other models in terms of mean AUC. For every single class, our proposed framework is better than all other models except with DensNet-121 for Fibrosis, Hernia, Mass, Nodule, Pneumonia, and Pneumothorax. Possible reasons can be that Rajpurkar et al’s backbone is much deeper than our ResNet-50\cite{rajpurkar2017chexnet}, which enables it to capture more discriminative features than our ResNet-50. In addition, “Mass” and “Nodule” parts are small and hard to detect. For “Fibrosis” and “Hernia,” they are not annotated with bounding boxes and diffuse, and thus we cannot apply the weakly-supervised learning with radiomic features.
%We also report results of ChesxRadiNet with ResNet-18, a relevant small network. We find that our model still performs better than other models except Rajpurkar et al\cite{rajpurkar2017chexnet}. It indicates the superior of our proposed method for using radiomic features.

\begin{table}[h]
\centering
\caption{AUC results on the NIH Chest X-ray dataset.}
\vspace{-1em}
\label{tab:disease id}
\begin{tabular}{lccccc}
\toprule
Method & Atelectasis & Cardiomegaly & Consolidation & Edema & Effusion \\
\midrule
Wang et al., 2017\cite{wang2017chestxray8} & 0.716 & 0.807 & 0.708 & 0.835 & 0.784 \\
Wang et al., 2018\cite{wang2018tienet} & 0.732 & 0.844 & 0.701 & 0.829 & 0.793 \\
Yao et al., 2018\cite{yao2017learning} & 0.772 & 0.904 & 0.788 & 0.882 & 0.859 \\
Rajpurkar et al., 2017\cite{rajpurkar2017chexnet} & 0.821 & 0.905 & 0.794 & 0.893 & 0.883 \\
Kumar et al., 2017\cite{kumar2017boosted} & 0.762 & 0.913 & 0.784 & 0.888 & 0.864 \\
ChexRadiNet & \textbf{0.831} & \textbf{0.934} & \textbf{0.817} & \textbf{0.906} & \textbf{0.892} \\
\midrule
Method & Emphysema & Fibrosis & Hernia & Infiltration & Mass \\
\midrule
Wang et al., 2017\cite{wang2017chestxray8} & 0.815 & 0.769 & 0.767 & 0.609 & 0.706 \\
Wang et al., 2018\cite{wang2018tienet} & 0.865 & 0.796 & 0.876 & 0.666 & 0.725 \\
Yao et al., 2018\cite{yao2017learning} & 0.829 & 0.767 & 0.914 & 0.695 & 0.792 \\
Rajpurkar et al., 2017\cite{rajpurkar2017chexnet} & \textbf{0.926} & \textbf{0.804} & \textbf{0.939} & 0.720 & \textbf{0.862} \\
Kumar et al., 2017\cite{kumar2017boosted} & 0.898 & 0.756 & 0.802 & 0.692 & 0.750 \\
ChexRadiNet & 0.925 & 0.798 & 0.882 & \textbf{0.734} & 0.846 \\
\midrule
Method & Nodule & Pleural Thickening & Pneumonia & Pneumothorax & \textbf{Mean} \\
\midrule
Wang et al., 2017\cite{wang2017chestxray8} & 0.671 & 0.708 & 0.633 & 0.806 & 0.738 \\
Wang et al., 2018\cite{wang2018tienet} & 0.685 & 0.735 & 0.720 & 0.847 & 0.772 \\
Yao et al., 2018\cite{yao2017learning} & 0.717 & 0.765 & 0.713 & 0.841 & 0.803 \\
Rajpurkar et al., 2017\cite{rajpurkar2017chexnet} & \textbf{0.777} & 0.814 & \textbf{0.763} & \textbf{0.893} & 0.842 \\
Kumar et al., 2017\cite{kumar2017boosted} & 0.666 & 0.774 & 0.715 & 0.859 & 0.795 \\
ChexRadiNet & 0.748 & \textbf{0.867} & 0.737 & 0.889 & \textbf{0.843}\\
\bottomrule
\end{tabular}
\end{table}

\subsubsection{Disease localization}

We compare our disease localization accuracy under varying IoU to other state-of-the-art models, shown in Table~\ref{tab:disease loc}. Our model predicts well not only for easy tasks but also for hard tasks like localizing “Mass” and “Nodule”, where the disease localization is within a small area. When the IoU is set to 0.1, our model outperforms other models in terms of Atelectasis, Cardiomegaly, Effusion, and Pneumothorax. As the IoU threshold increases, our framework is superior to other models in terms of better accuracy and maintains great performance. For instance, when IoU is set to 0.3, our result for “Cardiomegaly” is 0.73 while the reference model is only 0.46. We get more than 0.15 accuracy improvement for Effusion, Infiltration, Mass, Pneumonia, and Pneumothorax. When IoU is set to 0.5, our result for “Cardiomegaly” is still as high as 0.59 while the reference model drops to barely 0.18.

Following Li et al.\cite{li2018thoracic}, we prefer a higher IoU threshold, i.e., IoU = 0.7, for disease localization because we expect high-accuracy disease localization application in clinical use. To this end, the method we proposed is superior to the baseline by a large margin.

Please note that for some diseases, e.g., Pneumonia and Infiltration, the localization of disease can appear in multiple places while only one bounding box is provided for each image. Thus, it is reasonable that our model doesn’t align well with the ground truth when the threshold is as small as 0.1, especially for Pneumonia and infiltration.
Overall, our model outperforms the reference models for all IoU thresholds except for T(IoU)=0.1 (probably because ground truth has missing annotation while ours does not).

\begin{table}
\centering
\caption{Disease localization under varying IoU on the NIH Chest X-ray dataset. Please note that since our model doesn’t use any ground truth bounding box information, to fairly evaluate the performance of our model, we only consider the previous methods’ results under the same setting, therefore, for the case T(IoU)=0.1, we have two baselines, but for the rest cases, we only have one baseline.}
\vspace{-1em}
\label{tab:disease loc}
\small
\begin{tabularx}{\textwidth}{@{}cX@{~}c@{~~}c@{~~}c@{~~}c@{~~}c@{~~}c@{~~}c@{~~}c@{~~}c@{}}
\toprule
T(IoU) & Model & Atelectasis & Cardiomegaly & Effusion & Infiltration & Mass & Nodule & Pneumonia & Pneumothorax & \textbf{Mean} \\
\midrule
0.1 & Wang et al., 2017\cite{wang2017chestxray8} & 0.69 & 0.94 & 0.66 & 0.71 & 0.40 & 0.14 & 0.63 & 0.38 & 0.569 \\
 & Li et al., 2018\cite{li2018thoracic} & 0.63 & 0.89 & 0.78 & 0.91 & 0.70 & 0.29 & 0.31 & 0.44 & 0.619 \\
 & ChexRadiNet & 0.72 & 0.96 & 0.81 & 0.88 & 0.67 & 0.33 & 0.59 & 0.47 & \textbf{0.679}   \\
\midrule
0.2 & Wang et al., 2017\cite{wang2017chestxray8} & 0.47 & 0.68 & 0.45 & 0.48 & 0.26 & 0.05 & 0.35 & 0.23 & 0.371 \\
 & ChexRadiNet & 0.49 & 0.84 & 0.62 & 0.54 & 0.46 & 0.21 & 0.43 & 0.39 & \textbf{0.498 }\\
\midrule
0.3 & Wang et al., 2017\cite{wang2017chestxray8} & 0.24 & 0.46 & 0.30 & 0.28 & 0.15 & 0.04 & 0.17 & 0.13 & 0.221 \\
 & ChexRadiNet & 0.28 & 0.73 & 0.54 & 0.43 & 0.38 & 0.15 & 0.35 & 0.32 & \textbf{0.398} \\
\midrule
0.4 & Wang et al., 2017\cite{wang2017chestxray8} & 0.09 & 0.28 & 0.20 & 0.12 & 0.07 & 0.01 & 0.08 & 0.07 & 0.115 \\
 & ChexRadiNet & 0.17 & 0.65 & 0.42 & 0.32 & 0.29 & 0.09 & 0.21 & 0.19 & \textbf{0.293}   \\
\midrule
0.5 & Wang et al., 2017\cite{wang2017chestxray8} & 0.05 & 0.18 & 0.11 & 0.07 & 0.01 & 0.01 & 0.03 & 0.03 & 0.061 \\
 & ChexRadiNet & 0.11 & 0.59 & 0.29 & 0.15 & 0.12 & 0.07 & 0.14 & 0.08 & \textbf{0.194}   \\
\midrule
0.6 & Wang et al., 2017\cite{wang2017chestxray8} & 0.02 & 0.08 & 0.05 & 0.02 & 0.00 & 0.01 & 0.02 & 0.03 & 0.029 \\
 & ChexRadiNet & 0.06 & 0.37 & 0.09 & 0.06 & 0.08 & 0.04 & 0.05 & 0.05 & \textbf{0.100}   \\
\midrule
0.7 & Wang et al., 2017\cite{wang2017chestxray8} & 0.01 & 0.03 & 0.02 & 0.00 & 0.00 & 0.00 & 0.01 & 0.02 & 0.011 \\
 & ChexRadiNet & 0.02 & 0.21 & 0.04 & 0.02 & 0.07 & 0.01 & 0.03 & 0.04 & \textbf{0.055}  \\
\bottomrule
\end{tabularx}
\end{table}

\section{Discussion}

\subsection{Ablation study}

We conducted an ablation study to demonstrate the performance of radiomics on NIH Chest X-ray (Table~\ref{tab:comp nih}), CheXpert (Table~\ref{tab:comp chexpert}), and MIMIC-CXR (Table~\ref{tab:comp mimic}). We tried ResNet50+Triplet Attention without radiomic features. Table~\ref{tab:comp nih} shows that AUC will drop significantly when not using radiomic features. We observe the same trend in the other two datasets. This demonstrates that it is beneficial to include radiomic features.

We also report results of ChesxRadiNet using ResNet-18, a relevant small network, as a backbone. Table~\ref{tab:comp resnet18} shows the results with and without using the radiomic features in three datasets. We observe the AUCs drop significantly when not using radiomic features in all cases. This suggests that the generalizability of our proposed method in smaller networks. In addition, the ResNet-18 version still performs better than other models in Table~\ref{tab:disease id} except Rajpurkar et al\cite{rajpurkar2017chexnet}. It indicates the superior of our proposed method for using radiomic features.

\begin{table}[h]
\centering
\vspace{1em}
\caption{Comparison of AUC on the NIH Chest X-ray dataset.}
\label{tab:comp nih}
\vspace{-1em}
\begin{tabular}{lccccc}
\toprule
Method & Atelectasis & Cardiomegaly & Consolidation & Edema & Effusion \\
%\midrule
w/o radiomics & 0.751 & 0.850 & 0.777 & 0.867 & 0.833 \\
ChexRadiNet & 0.831 & 0.934 & 0.817 & 0.906 & 0.892 \\
\midrule
Method & Emphysema & Fibrosis & Hernia & Infiltration & Mass \\
%\midrule
w/o radiomics & 0.783 & 0.733 & 0.804 & 0.670 & 0.694 \\
ChexRadiNet & 0.925 & 0.798 & 0.882 & 0.734 & 0.846 \\
\midrule
Method & Nodule & Pleural   Thickening & Pneumonia & Pneumothorax & \textbf{Mean} \\
%\midrule
w/o radiomics & 0.643 & 0.699 & 0.700 & 0.792 & 0.757 \\
ChexRadiNet & 0.748 & 0.867 & 0.737 & 0.889 & \textbf{0.842}\\
\bottomrule
\end{tabular}
\vspace*{1em}
\end{table}

\begin{table}[h]
\centering
\caption{Comparison of AUC on the CheXpert dataset.}
\vspace{-1em}
\label{tab:comp chexpert}
\begin{tabular}{lcccccc}
\toprule
Method & Atelectasis & Cardiomegaly & Consolidation & Edema & Pleural Effusion & \textbf{Mean}\\
\midrule
w/o radiomics & 0.781 & 0.813 & 0.893 & 0.918 & 0.921 & 0.865\\
ChexRadiNet & 0.831 & 0.848 & 0.920 & 0.930 & 0.921 & \textbf{0.890}\\
\bottomrule
\end{tabular}
\end{table}
\begin{table}[H]
\centering
%\vspace{1em}
\caption{Comparison of AUC on the MIMIC-CXR dataset.}
\vspace{-1em}
\label{tab:comp mimic}
\begin{tabular}{lccccc}
\toprule
Method & Atelectasis & Cardiomegaly & Consolidation & Edema & Enlarged Card. \\
%&&&&&Cardiomediastinum\\
%\midrule
w/o radiomics & 0.841 & 0.824 & 0.859 & 0.906 & 0.748 \\
ChexRadiNet & 0.851 & 0.831 & 0.866 & 0.900 & 0.767 \\
\midrule
Method & Fracture & Lung Lesion & Lung Opacity & Pleural Effusion& Pneumonia \\
%&&&&Effusion\\
%\midrule
w/o radiomics & 0.713 & 0.782 & 0.775 & 0.923 & 0.753 \\
ChexRadiNet & 0.735 & 0.814 & 0.810 & 0.933 & 0.831 \\
\midrule
Method & Pneumothorax & Pleural Other & Support Devices & \textbf{Mean} &  \\
%\midrule
w/o radiomics & 0.909 & 0.850 & 0.931 & 0.832 &  \\
ChexRadiNet & 0.919 & 0.909 & 0.937 & \textbf{0.854} & \\
\bottomrule
\end{tabular}
\end{table}

\begin{table}[h]
\centering
%\vspace{1em}
\caption{Comparison of mean AUC on three datasets using ResNet-18 as a backbone.}
\vspace{-1em}
\label{tab:comp resnet18}
\begin{tabular}{lccc}
\toprule
& NIH Chest X-ray & CheXpert & MIMIC-CXR\\
\midrule
w/o radiomics & 0.749 & 0.854 & 0.822\\
ChesxRadiNet (ResNet-18) & 0.810 & 0.883 & 0.837\\
\bottomrule
\end{tabular}
\end{table}

\subsection{Qualitative analysis}

\begin{figure}[h]
\centering
\includegraphics[width=\textwidth]{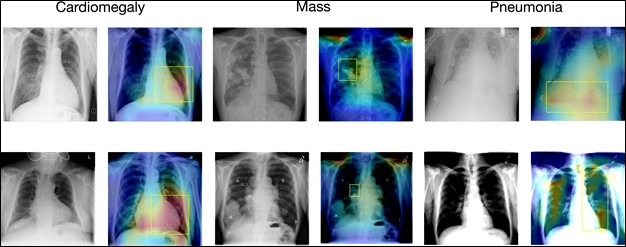}
\caption{Visualization of the disease localization on the test images with ChexRadiNet and ground truth bounding boxes. The attention maps are generated from the final output tensor and overlapped on the original radiology images. The left image in each pair is the chest X-ray image and the right one is the generated attention map and the ground truth (in the yellow box).}
\label{fig:map}
\vspace{1em}
\end{figure}

Figure~\ref{fig:map} shows the attention map of our model against the ground truth bounding boxes. The visualization provides better explainability of our model. In Figure~\ref{fig:map} we visualized our results for Cardiomegaly, Mass, and Pneumonia.

Cardiomegaly is considered to be present if the cardiothoracic rate is larger than 50\% (cardiothoracic Ratio equals “Maximum horizontal cardiac width” over “Maximum horizontal thoracic width”), which means an enlarged heart. The 2nd image in the 1st row as well as the 2nd image in the 2nd row in Figure~\ref{fig:map} shows that our model successfully detects cardiomegaly, an enlarged heart, perfectly, and aligns with the yellow bounding box well.

A lung mass is an abnormal spot in the lungs that is more than 3 centimeters. Our results (4th images in the 1st and 2nd rows), although focusing on larger areas, can capture some clues of lung mass.

Note that in the chest X-ray 14 dataset, only one bounding box is annotated for one disease image. Though some patients are diagnosed with several diseases, only the most important disease is annotated on the radiology image. This means that ground truth has missing annotations (shown by Pneumonia). Pneumonia inflames the air sacs in one or both lungs. For Pneumonia detection, radiologists will look for white spots in the lungs. For the 6th image in the 2nd row, both lungs are infected and white spots are shown in both lungs. However, the bounding box of the 6th image only annotates the right lung while our model successfully localizes Pneumonia for both lungs.

Overall, our results show that the predicted disease localizations have a great alignment with the ground truth and can even serve as a supplement to the ground truth.

\section{Conclusion}

We propose a framework that jointly learns radiomic features and predicts 14 thoracic diseases. We evaluated our model on three publicly available corpora. We showed that both our disease identification and localization outperform state-of-the-art models in the quantitative and qualitative analysis. 

Our proposed framework has two main limitations. First, chest X-rays are very different from natural images, but we rely on deep learning models (ResNet) that work better on natural images. Second, the robustness of radiomic features relies on the accuracy of bounding boxes, in our work, the bounding boxes are generated by heatmaps. It is not guaranteed that the generated heatmaps are always good and accurate. Our future work will continue to solve these two limitations.

Automatically generating correct bounding boxes can be a milestone to push the agenda for AI-driven medical imaging diagnosis. It can abruptly increase the annotated medical images at a much lower cost so that better CNN models can be trained, therefore better diagnosis models can be obtained. Bounding boxes can increase the interpretability of AI solutions by locating the abnormalities as the visual evidence in medical images, which can build trust between doctors and patients.

\section*{Acknowledgment}

This work is supported by Amazon Machine Learning Research Award 2020. It also was supported by the National Library of Medicine under Award No. 4R00LM013001.

\makeatletter
\renewcommand{\@biblabel}[1]{\hfill #1.}
\makeatother

\bibliography{ref}
\bibliographystyle{unsrt}
%\begin{thebibliography}{1}

%\bibitem{ref1}
%Pryor TA, Gardner RM, Clayton RD, Warner HR. The HELP system. J Med Sys. 1983;7:87-101.
%\bibitem{ref2}
%Gardner RM, Golubjatnikov OK, Laub RM, Jacobson JT, Evans RS. Computer-critiqued blood ordering using the HELP system. Comput Biomed Res 1990;23:514-28.

%\end{thebibliography}

\end{document}